\documentclass{article}
\usepackage{spconf,amsmath,graphicx}
\usepackage{enumerate}

\title{Nose, Eyes and Ears:  Head Pose Estimation by Locating Facial Keypoints}
\name{Aryaman Gupta, Kalpit Thakkar, Vineet Gandhi and P J Narayanan}
\address{Centre for Visual Information Technology, KCIS, IIIT Hyderabad}

\begin{document}

\maketitle

\begin{abstract}
Monocular head pose estimation requires learning a model that computes the intrinsic Euler angles for pose (yaw, pitch, roll) from an input image of human face. Annotating ground truth head pose angles for images in the wild is difficult and requires ad-hoc fitting procedures (which provides only coarse and approximate annotations). This highlights the need for approaches which can train on data captured in controlled environment and generalize on the images in the wild (with varying appearance and illumination of the face). Most present day deep learning approaches which learn a regression function directly on the input images fail to do so. To this end, we propose to use a higher level representation to regress the head pose while using deep learning architectures. More specifically, we use the uncertainty maps in the form of 2D soft localization heatmap images over five facial keypoints, namely left ear, right ear, left eye, right eye and nose, and pass them through an convolutional neural network to regress the head-pose. We show head pose estimation results on two challenging benchmarks BIWI and AFLW and our approach surpasses the state of the art on both the datasets. 
\begin{keywords}
Image analysis, Pose estimation
\end{keywords} \vspace{-1em}
\end{abstract}

\section{Introduction}
The ability of humans to comprehend non-verbal communication by effortlessly estimating the orientation and movements of human head is fascinating. In order to humanize machines by bringing them closer to human-like perception and understanding, accurately estimating the human head orientation using visual imagery presents an important challenge. Head pose relates to the visual attention and interest of a person, which is crucial for many applications in computer vision. Estimating head pose has been actively pursued in problems like social event analysis \cite{varadarajan2018joint}, Human Computer Interaction (HCI) \cite{Wang:2018:FG}, driver assistance systems \cite{schwarz2017driveahead} etc., which are an important part of present day technologies.

\begin{figure*}[t]
    \centering
    \begin{tabular}{ccccccc}
        \resizebox{0.12\textwidth}{0.12\textwidth}{\includegraphics{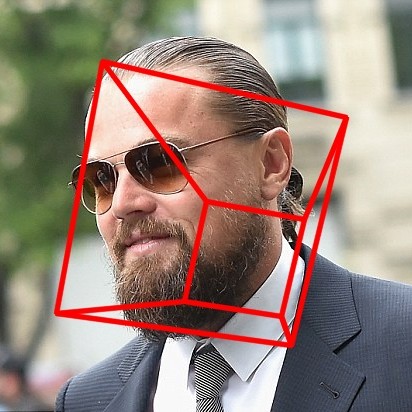}} & 
        \resizebox{0.12\textwidth}{0.12\textwidth}{\includegraphics{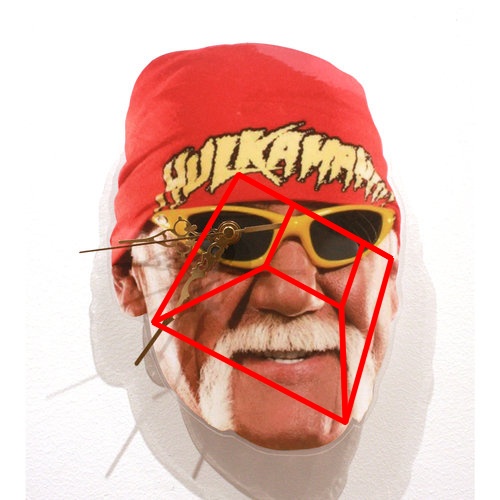}} &
        \resizebox{0.12\textwidth}{0.12\textwidth}{\includegraphics{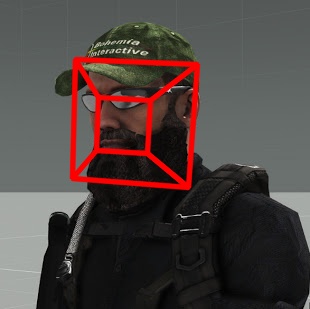}} &
        \resizebox{0.12\textwidth}{0.12\textwidth}{\includegraphics{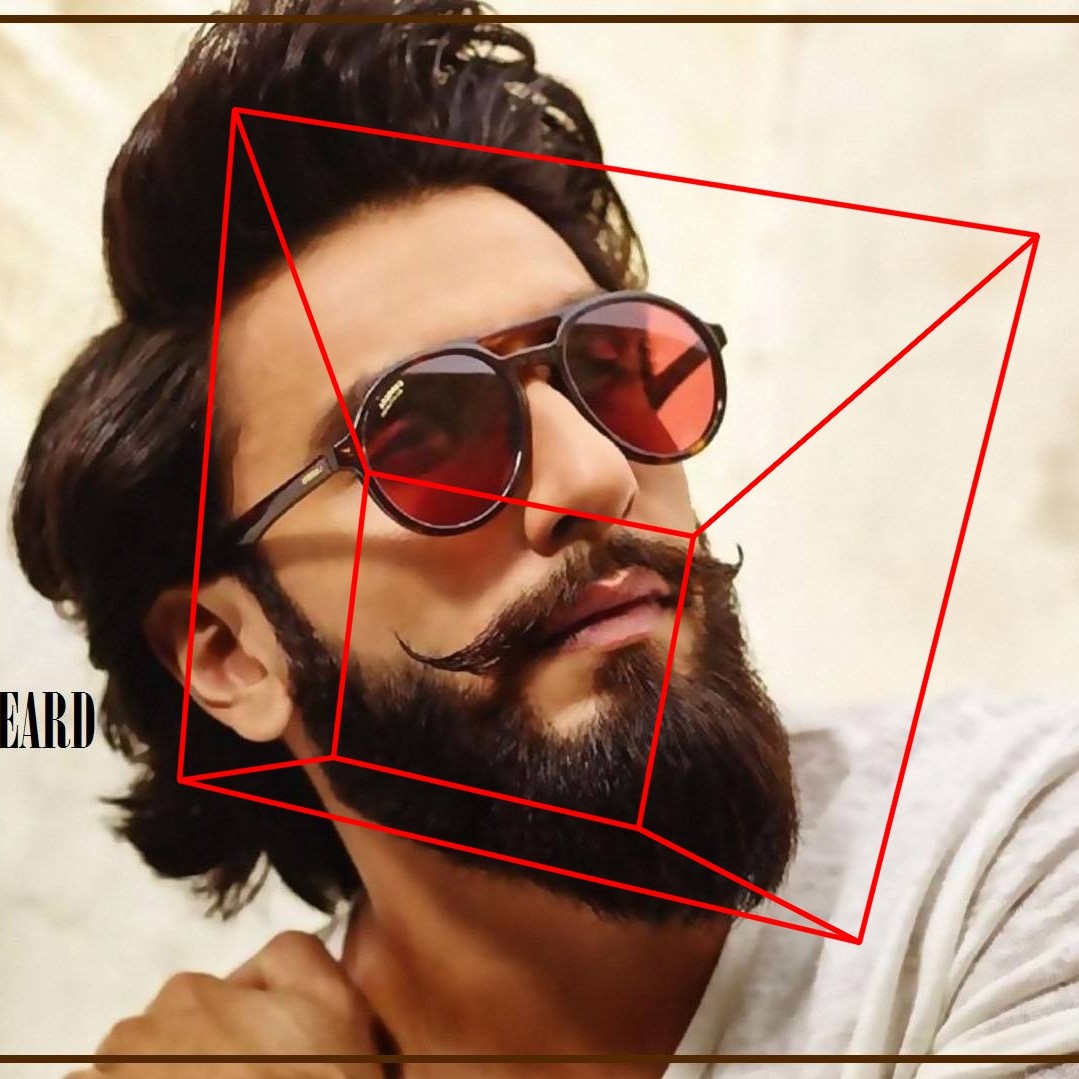}} &
        \resizebox{0.12\textwidth}{0.12\textwidth}{\includegraphics{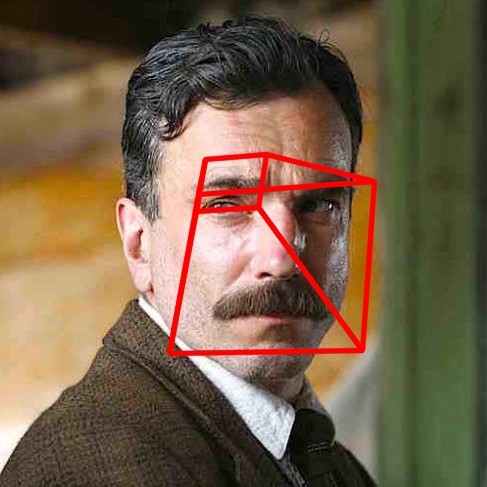}} &
        \resizebox{0.12\textwidth}{0.12\textwidth}{\includegraphics{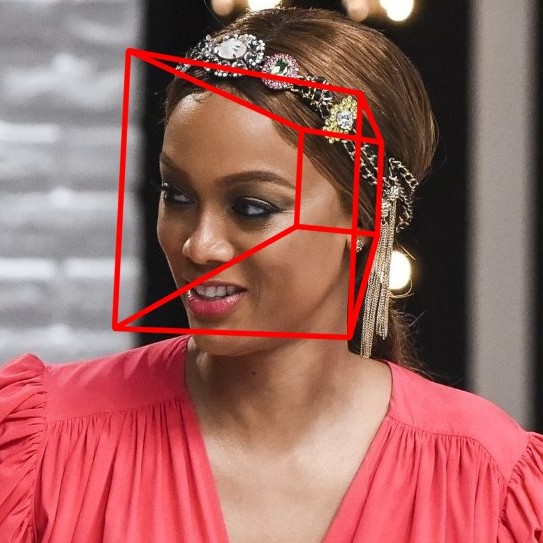}} \\
        \resizebox{0.12\textwidth}{0.12\textwidth}{\includegraphics{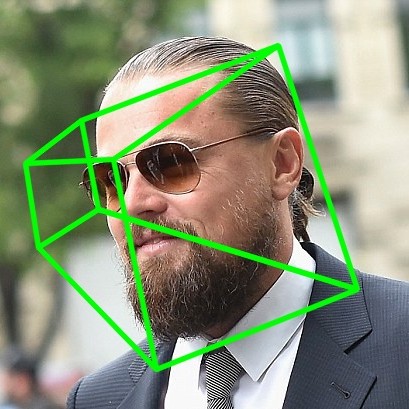}} &
        \resizebox{0.12\textwidth}{0.12\textwidth}{\includegraphics{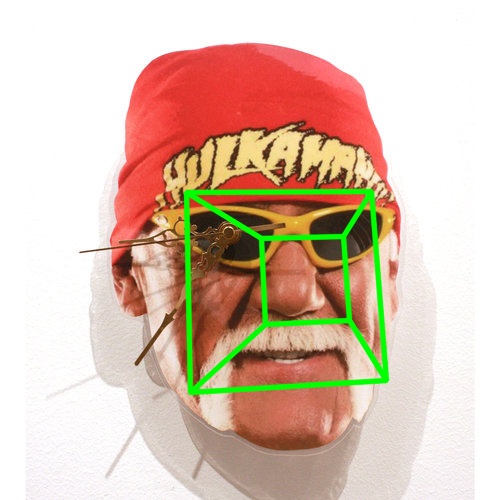}} &
        \resizebox{0.12\textwidth}{0.12\textwidth}{\includegraphics{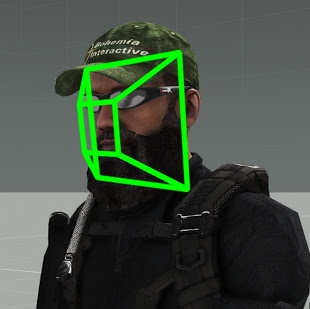}} &
        \resizebox{0.12\textwidth}{0.12\textwidth}{\includegraphics{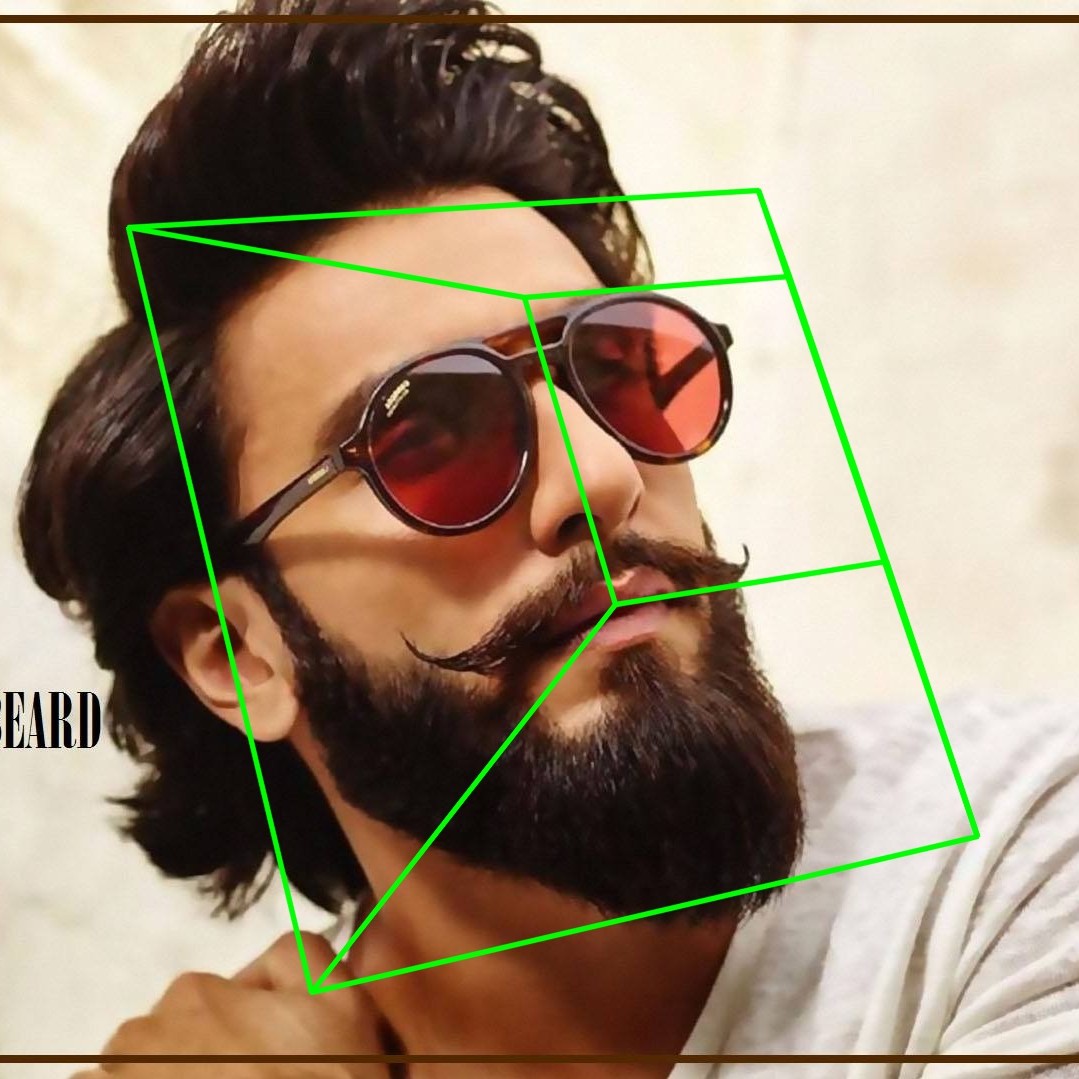}} &
        \resizebox{0.12\textwidth}{0.12\textwidth}{\includegraphics{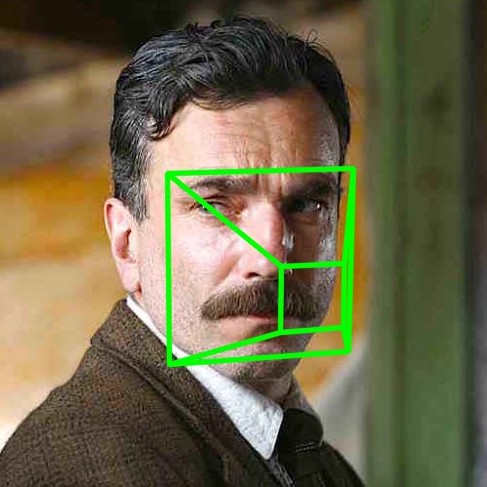}} &
        \resizebox{0.12\textwidth}{0.12\textwidth}{\includegraphics{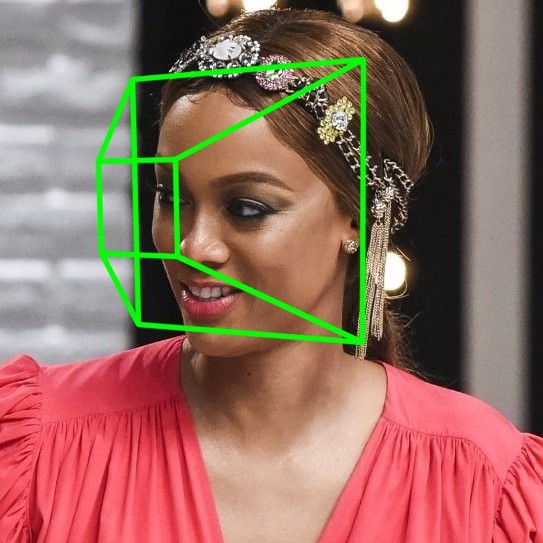}}
    \end{tabular}
    \caption{Estimation of head pose using three different models (all trained on BIWI), on unseen images taken from the web. \textbf{Top row}: Results for CNN-based model \cite{BMVC2015_130} which takes RGB images as input, \textbf{Bottom row}: Results for our CNN-based framework which takes heatmaps of five facial keypoints locations as input.}
    \label{fig:teaser}
\end{figure*}
Formally, head pose estimation entails computing the 3D orientation of head with respect to the camera pose using digital images. Initial approaches estimated only one or two angles for head pose while assuming other angles are fixed or fixed discrete values for head pose angles to be estimated \cite{Huang1998svmfacepose,Kwong2002CompositeSV}. However, head pose estimation with \textit{three} degrees of freedom, viz.\ (yaw, pitch and roll), is more useful than discrete head pose and recent methods have been aimed at estimating the three head pose angles. With the availability of well annotated datasets captured using Kinect sensors such as BIWI \cite{eth_biwi_00839}, monocular head pose estimation with 3-DOF has seen good improvements in recent years. The state-of-the-art method relies on end-to-end convolutional regression networks \cite{lathuiliere2017invreg}, which takes RGB images as input and learns the parameters of an inverse regression network using a Mean Squared Error (MSE) loss. As BIWI \cite{eth_biwi_00839} is captured in a controlled environment for accurate ground truth annotation which is dependent on precise 3D reconstruction of face, methods using RGB input directly for head pose estimation on BIWI \cite{eth_biwi_00839} fail to generalize on images in the wild (as illustrated in Figure~\ref{fig:teaser}). On the other hand, datasets like AFLW \cite{koestinger11a} only provide coarse approximation of ground truth angles as annotation of ground truth on images in the wild is challenging. Hence, an important property for head pose estimation algorithms is generalization on face images in the wild when trained on precisely annotated datasets like BIWI \cite{eth_biwi_00839}.

While computer vision based pose estimation approaches have focused predominantly on appearance-based solutions that compute human pose directly from digital images, there have been methods based on psychophysical experiments. These consider the human perception of head pose to rely on cues such as deviation of nose angle and the deviation of the head from bilateral symmetry \cite{WILSON2000459}. Since it is easier to annotate 2D keypoints directly on images, huge labelled datasets are now available \cite{andriluka14cvpr} and have lead to development of powerful methods \cite{cao2017realtime} for localizing keypoints like nose, eyes and ears. We hypothesize that we can learn a head pose estimation model using only five facial keypoint locations. Such a model implicates an abstraction over the appearance and illumination dependent image data which is a hindrance for generalization capability of head pose estimation methods. The abstraction limits the dependencies of the model to scale and configuration of a few keypoint locations.

Our first baseline approach takes as input the keypoint locations and directly predicts the head-pose using a Multi Layer Perceptron (MLP). However, we notice that the facial keypoint locations have inherent uncertainty in their estimation. Hence we propose a second framework, which first computes the uncertainty maps for the five points in the form of heatmap images capturing their soft localization (in other words, the probability distribution of all possible locations of that keypoint). The five images are then stacked together and provided as input to a Convolutional Neural Network (CNN) for estimation of head pose angles. We show that our baseline approach achieves competitive performance, while CNN-based framework surpasses state-of-the-art. The contributions of this paper are as follows:
\begin{itemize}
    \item A hypothesis on learning a model for head pose estimation which relies only on five facial keypoint locations and abstracts out the dependency on appearance of the subject. 
    \item A baseline approach that uses the exact keypoint locations (sampled from their distribution) and employs a MLP for regression of pose angles. 
    \item A CNN-based framework which uses the probability distribution of keypoint locations in the form of heatmap images, as input to regress the head pose.
    \item State-of-the-art performance for head pose estimation using the CNN-based framework on the BIWI \cite{eth_biwi_00839} and AFLW \cite{koestinger11a} datasets.
\end{itemize}

\section{Related Work}
\label{sec:relwork}
Previous approaches to head pose estimation can be classified into two categories: RGB and RGBD based (2D vs 3D input). We limit our discussion to RGB input only. Earlier methods for head pose estimation used appearance templates that use a set of exemplars to find the pose of an input image, by finding the closest exemplar \cite{Huang1998svmfacepose}. The assumption that similarity in image space equates similarity in pose is the major drawback of such methods. Extending appearance templates, several methods using multiple pose detectors (each corresponding to one discrete pose) have been proposed \cite{Kwong2002CompositeSV}. However, detector-based methods require several detectors and non-face samples (negative samples) for successful training, which is burdensome. Manifold embedding methods were later introduced, which project an input sample to a lower dimension using an embedding function and regress pose in the embedding space. Techniques like PCA \cite{Wu:2008:THP}, Isomap \cite{Hu2005nonlinear} and several combinations \cite{yan20083dpose} of dimensionality reduction approaches are used for head pose estimation. Learning useful low-dimensional representations needs proper training data having balanced samples. 

With the transition to deep learning based methods, several former drawbacks have been mitigated. One of the earliest efforts in this area was by Osadchy et. al \cite{Osadchy:2007:SFD}. They extract CNN features from images and regress pose using them. Patacchiola and Cangelosi \cite{Patacchiola2017HeadPE} test the effect of dropout and adaptive gradient-based methods combined with CNNs for head pose estimation, where they propose to use adaptive gradients in conjunction with a CNN. On the other hand, Ruiz et. al \cite{ruiz2017nokeypoints} propose a CNN with $3$ separate branches, each with combined classification and regression for the respective head pose angle. Both these methods aim to improve performance of head pose estimation in the wild. Lathuili\'ere et. al \cite{lathuiliere2017invreg} proposed a CNN-based model with a Gaussian mixture of linear inverse regressions. They use an Imagenet-pretrained CNN to learn face features and train a pose regressor on them. An extension of this approach by Drouard et. al \cite{drouard:hal-01413406} proposes to cope with changing illumination conditions, variability in face orientation and in appearance, etc. by combining the qualities of unsupervised manifold learning and inverse regressions. However, as the CNN-based methods estimate the pose angles directly from RGB images, it makes them prone to poor generalization on account of illumination ass well as appearance changes.
\begin{figure*}[t]
    \begin{center}
        \begin{tabular}{ccccccc}
            \includegraphics[scale=0.55]{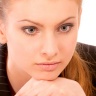} &
            \includegraphics[scale=0.55]{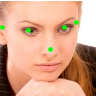}  &
            \includegraphics[scale=0.55]{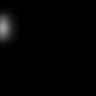} &
            \includegraphics[scale=0.55]{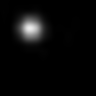} &
            \includegraphics[scale=0.55]{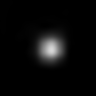} &
            \includegraphics[scale=0.55]{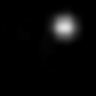}&
            \includegraphics[scale=0.55]{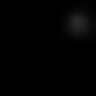} \\
            Input image &
            keypoints &
            Left Ear &
            Left Eye &
            Nose & 
            Right Eye &
            Right Ear
        \end{tabular}
    \end{center} \vspace{-1.2em}
    \caption{Example of a face image, detected keypoints and respective heatmaps of each keypoint computed using \cite{cao2017realtime}.}
    \label{fig:im2hm}
\end{figure*}
Geometric models regress the pose using facial features such as keypoints, nose angle, etc. and have been proposed in previous literature \cite{Wang:2007:EM3Dheadpose}. Similar in spirit, we propose to use a higher-level feature to drive the pose regression, viz. the heatmaps of five facial keypoints extracted from face images (or exact 2D locations) using a keypoint localization routine \cite{cao2017realtime}. The performance of our models prove our hypothesis of facilitating abstraction over illumination and appearance dependent image data by achieving state-of-the-art results for head pose estimation and demonstrating good generalization capability.

\section{Head Pose Estimation via keypoint localization}
\label{sec:cnnmodel}
Our baseline approach is to employ a Multi Layer Perceptron (MLP) which regresses the 3D head-pose directly using the predicted locations of the five keypoints (detected using \cite{cao2017realtime}). Each of the keypoint is parameterized by its 2D location and prediction likelihood, resulting in an input vector of 15 dimensions, which is used to regress a 3D vector representing the yaw, pitch and roll. Undetected keypoints are represented by a vector of zeroes.

MLP-based method is based on the assumption that the locations of five facial keypoints estimated from the face image are accurate. However, in practice there is inherent uncertainty in predicting the locations of keypoints such as eyes, ear and nose, using an optimization based approach \cite{cao2017realtime}. One possible way to account for this uncertainty in localization is to treat the image locations of the facial keypoints as latent variables. From a representation perspective, uncertainty maps (heatmap images) can be used to depict latent variables, which capture the soft localization of 2D keypoint locations (Figure \ref{fig:im2hm} illustrates an image and corresponding uncertainty maps for the five different facial keypoints used in our work). An image-based representation of the facial keypoint locations facilitates the use of CNN-based approaches for learning the head pose.
\begin{figure}
    \begin{center}
        \includegraphics[ height=3.2cm, width=8.5cm]{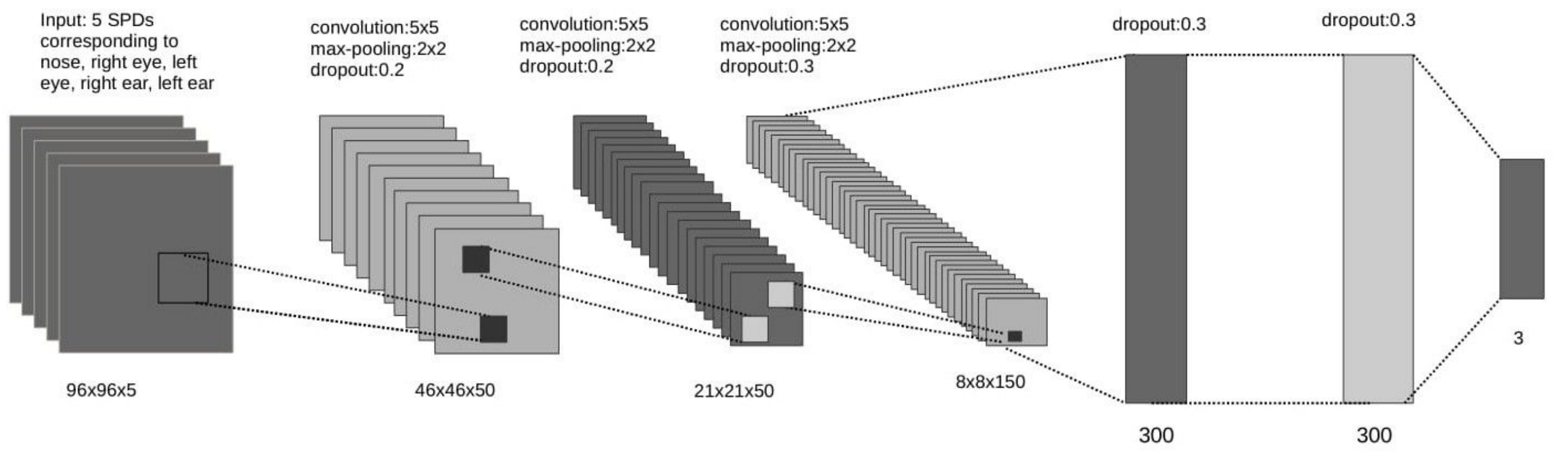}
    \end{center} \vspace{-1.3em}
    \caption{The architecture consists of $3$ convolutional layers (conv1, conv2, conv3) followed by two fully connected layers (fc1, fc2). The input has $5$ channels: one each for the nose, left eye, right eye, left ear and right ear (heatmap images for these keypoints). The network outputs the estimated values of the three intrinsic Euler angles (yaw, pitch, roll).}
    \label{fig:cnnArch}
\end{figure}
Uncertainty maps over locations of keypoints (or joints) in human body or an object skeleton, present in an image, have been successfully used in previous literature where the exact locations of the keypoints were noisy or unknown. Zhou \cite{zhou2016sparseness} use heatmap images of 2D joint locations to infer 3D human pose using an Expectation Maximization framework. Wu \cite{wu2016single} use heatmaps of 2D skeleton keypoints of an object as an intermediate representation to recover 3D structure of an object and bridge the gap between synthetic and real data. Interestingly, both these works \cite{zhou2016sparseness,wu2016single} use heatmaps over 2D spatial locations to infer 3D structure/pose. Deriving motivation from these efforts, we propose an algorithm which takes 2D uncertainty maps over the facial keypoints as input and regresses the 3D head pose. 

Unlike previous efforts \cite{zhou2016sparseness,wu2016single} that use heatmaps as an intermediate representation and do not have ground truth data, we have ground truth pose angles available. This allows us to directly train a convolutional regression network using ground truth supervision for head pose estimation. Specifically, we use OpenPose \cite{cao2017realtime} to compute the uncertainty maps for the five facial keypoint locations as illustrated in Figure \ref{fig:im2hm}. Each heatmap image is considered as a separate channel and the channels are stacked together, which generates a 5-channel feature map. This feature map is used as an input to the CNN, the architecture of which is shown in Figure \ref{fig:cnnArch}, to learn a head pose estimation model. The final layer gives the values of three pose angles obtained as a result of the convolutional regression. We use a MSE loss to train the convolutional regression network, which can be written as follows:
\begin{align}
    \mathbf{L}_{\text{mse}} &= \frac{1}{3} \sum_{i=1}^3 \big(\mathbf{\Theta}_i - \mathbf{\hat{\Theta}}_i\big)^2
\end{align}
where, $\mathbf{\Theta}_i$ is the vector consisting of the predicted values for intrinsic Euler angles and $\mathbf{\hat{\Theta}}_i$ is the vector consisting of the values of ground truth angles.
\section{Experiments and Results}
\label{sec:expresults}

\subsection{Experimental Setup and Datasets}
\noindent \textbf{MLP-based Model}  Our network consists two hidden layers of size 30 neurons each. We set learning rate of $0.00001$ and train for $500$ epochs using Adam optimizer with a weight decay of $0.0001$ and batch size $64$. 

\vspace{0.5em}
\noindent \textbf{CNN-based Model} We use a CNN architecture with $3$ convolution layers and $2$ fully connected layers (we have used same architecture used in Liu\cite{7532566} but with 5 input channels).
Training is run for $1200$ epochs with Adam optimizer and set learning rate of $0.00001$. We set the batch size to $32$.
All the experiments are run on a single Nvidia GTX 1080Ti GPU. 
\vspace{0.5em}

\noindent We use two benchmark datasets to measure the performance of our models and test them. \textbf{BIWI} Kinect Headpose Dataset \cite{eth_biwi_00839} contains over 15K samples spread over 24 sequences, captured in a controlled environment. The range of head pose angles in the dataset vary from $\pm {75}^{\circ}$ for yaw, $\pm {60}^{\circ}$ for pitch and $\pm {50}^{\circ}$ for roll. \textbf{AFLW} \cite{koestinger11a} Annotated Facial Landmarks in the Wild (AFLW) provides a large-scale collection of annotated face images gathered from the web, exhibiting a large variety in appearance (e.g., pose, expression, ethnicity, age, gender) as well as general imaging and environmental conditions. In total about 25K faces are annotated with up to $21$ landmarks per image. 
\begin{table}[t]
    \begin{center}
        \begin{tabular}{|c|c|c|c|c|}
        \hline
            Method & Yaw & Pitch & Roll & MAE \\
        \hline    
            Liu \cite{7532566} & 6.0 & 6.1 & 5.7 & 5.94 \\
            Ruiz et al. \cite{ruiz2017nokeypoints} & 4.810 & 6.606 & 3.269 & 4.895 \\
            Drouard \cite{drouard:hal-01413406} & 4.24 & 5.43 & 4.13 & 4.6 \\
            DMLIR \cite{lathuiliere2017invreg} & \textbf{3.12} & 4.68 & 3.07 & 3.62 \\
            MLP with location (Ours) & 3.64 & 4.42 & 3.19 & 3.75 \\
            CNN + Heatmaps (Ours) & 3.46 & \textbf{3.49} & \textbf{2.74} & \textbf{3.23} \\
        \hline    
        \end{tabular}
    \end{center} \vspace{-0.5em}
    \caption{Results on BIWI with 8-fold cross-validation  (21 randomly selected videos for training and the remaining 3 videos for test such that no person appears both in training and test sets)}
    \label{tab:biwi}
\end{table}
\subsection{Results}
\label{subsec:res}

\noindent \textbf{Results on BIWI dataset:} As BIWI is captured in controlled conditions and has better ground truth annotations, better performance is achieved on this dataset. The motivation for designing our frameworks is to train a model on a dataset like BIWI and use it to generalize to face images in the wild. In order to demonstrate the ability of our frameworks, we predict the head pose on unseen images taken from the web (as illustrated in Figure \ref{fig:teaser}). Our results show the presence of a perceptually better sense of pose than a model learned directly on the RGB images. Quantitative results for the dataset in terms of Mean Absolute Error (MAE) from ground truth annotations are shown in Table \ref{tab:biwi} which shows that the MLP model achieves competitive performance, while the CNN based approach surpasses the state of the art.
\begin{table}
    \begin{center}
        \begin{tabular}{|c|c|c|c|c|}
        \hline
            Method & Yaw & Pitch & Roll & MAE \\
        \hline    
            View manifolds \cite{7301354} & -- & -- & -- & 17.52 \\
            Random Forests \cite{random_forest} & -- & -- & -- & 12.26 \\
            {Pata. and Cang.}$^{\ast}$ \cite{Patacchiola2017HeadPE} & 11.04 & 7.15 & 4.4 & 7.53 \\
            MLP + Locations (Ours)  & 9.56 & 6.64 & 4.68 & 6.96 \\
            CNN + Heatmaps (Ours) & \textbf{6.19} & \textbf{5.58} & \textbf{3.76} & \textbf{5.18} \\
        \hline    
        \end{tabular}
    \end{center} \vspace{-0.5em}
    \caption{Results on AFLW dataset with 5-fold cross validation. $^{\ast}$: Constrains the angles to a certain range. }
    \label{tab:aflw}
\end{table}

\begin{table}
    \begin{center}
        \begin{tabular}{|c|c|c|c|c|}
        \hline
            Method & Yaw & Pitch & Roll & MAE \\
        \hline    
            Kepler \cite{kepler} & 6.45 & 7.05 & 5.85 & 6.45 \\
            Ruiz et al. \cite{ruiz2017nokeypoints} & 6.26 & 5.89 & 3.82 & 5.324 \\
            MLP + Locations (Ours) & 6.02 & 5.84 & 3.56 & 5.14 \\
            CNN + Heatmaps (Ours) & \textbf{5.22} & \textbf{4.43} & \textbf{2.53} & \textbf{4.06} \\
        \hline    
        \end{tabular}
    \end{center} \vspace{-0.5em}
    \caption{Results on AFLW using testing protocol in \cite{kepler}.}
    \label{tab:aflw1}
\end{table}

\noindent \textbf{Results on AFLW dataset} Given the large variations in AFLW dataset, most of the previous methods compute results for head pose estimation on this dataset by constraining the range of angles, using a subsampled set of images or creating a very small test set \cite{ruiz2017nokeypoints,Patacchiola2017HeadPE}. We do not assume any such constraints and show the results using a standard five-fold validation process on the entire dataset, where the samples are randomly divided into train and test sets with 80\% samples ending up in training set. We also perform experiment following testing protocol in \cite{kepler} (i.e. selecting 1000 images from testing and remaining for training) and present the results in Table~\ref{tab:aflw1}. The numbers of other methods in both tables are reported directly from the associated papers (aligned with corresponding protocol). 

The results clearly show that our CNN-based framework achieves the lowest MAE, significantly improving on the previous state-of-the-art on both the protocols. Interestingly, the MLP based approach also gives competitive performance as compared to previous work. We believe that the exact locations of the facial keypoints, as used in case of MLP, makes it prone to overfitting while the heatmaps act as a regularizer in that sense, giving an edge to CNN based framework. Overall, the experiments provide a strong empirical evidence towards the hypothesis pursued in this paper.

\section{Conclusions}
\label{sec:conc}
In this paper, we present a hypothesis that using an intermediate representation such as locations of five facial keypoints instead of face images can help achieve better pose estimation and generalization performance. We propose two frameworks (a baseline approach employing MLP and a CNN over uncertainty maps) to support our claim. Although, minimal the MLP based approach gives competitive performance and we believe that it will improve with improvement in localization of keypoints. Owing to presence of noise in localization estimates, our CNN-based approach uses it as an advantage by representing the uncertainty as heatmaps and regressing the head pose with the heatmaps as input. The CNN-based framework surpasses state-of-the-art for head pose estimation on two challenging benchmarks BIWI \cite{eth_biwi_00839} and AFLW \cite{koestinger11a}. 

\bibliographystyle{IEEEbib}
\bibliography{headpose.bbl}

\end{document}